%% file: main.tex
\pdfoutput=1

\documentclass[11pt]{article}

\usepackage[]{ACL2024-mod}

\usepackage{times}
\usepackage{latexsym}

\usepackage[T1]{fontenc}

\usepackage[utf8]{inputenc}

\usepackage{microtype}

\usepackage{inconsolata}

\input{imp-def.tex}

\title{A Simple but Effective Approach to Improve Structured Language Model Output for Information Extraction}

\author{
  Yinghao Li, Rampi Ramprasad, Chao Zhang\\
  Georgia Institute of Technology, Atlanta, USA\\
  \texttt{$\{$yinghaoli,chaozhang$\}$@gatech.edu}\quad \texttt{rampi.ramprasad@mse.gatech.edu}
}

\begin{document}
\maketitle

\input{sections/0.abs.tex}

\input{sections/1.intro.tex}

\input{sections/8.related.tex}

\input{sections/4.method.tex}

\input{sections/5.results.tex}

\input{sections/6.conclusion.tex}

\input{sections/7.limit.tex}

\input{sections/9.ack.tex}

\bibliography{reference}
\bibliographystyle{acl_natbib}

\clearpage

\appendix

\input{appendix/a2.setup.tex}

\input{appendix/a3.results.tex}

\end{document}

%% file: imp-def.tex
\usepackage{times}
\usepackage{latexsym}
\usepackage{cite}
\usepackage{natbib}

\usepackage[T1]{fontenc}

\usepackage{microtype}

\usepackage{amsmath}
\usepackage{amsfonts}
\usepackage{graphicx}  
\usepackage{xspace}
\usepackage{eufrak}
\usepackage{bm}
\usepackage{multirow}
\usepackage{booktabs}
\usepackage{subfig}
\usepackage[utf8]{inputenc}

\usepackage{cleveref}
\usepackage{enumitem}
\usepackage{adjustbox}

\usepackage{xargs}
\usepackage{etoolbox}
\usepackage[flushleft]{threeparttable}

\usepackage[colorinlistoftodos,prependcaption,textsize=tiny]{todonotes}
\usepackage[group-separator={,}, group-minimum-digits=4]{siunitx}
\usepackage{comment}
\usepackage{listings}
\usepackage{xcolor}

\definecolor{codegreen}{rgb}{0,0.6,0}
\definecolor{codegray}{rgb}{0.5,0.5,0.5}
\definecolor{codepurple}{rgb}{0.58,0,0.82}
\definecolor{backcolour}{rgb}{0.95,0.95,0.95}

\lstdefinestyle{mystyle}{
    backgroundcolor=\color{backcolour},   
    commentstyle=\color{codegreen},
    keywordstyle=\color{magenta},
    numberstyle=\tiny\color{codegray},
    stringstyle=\color{codepurple},
    basicstyle=\tt\scriptsize,
    breakatwhitespace=false,         
    breaklines=true,                 
    captionpos=t,                    
    keepspaces=true,                 
    numbers=left,                    
    numbersep=5pt,                  
    showspaces=false,                
    showstringspaces=false,
    showtabs=false,                  
    tabsize=2,
}

\setlist{nosep}

\crefformat{section}{\S~#2#1#3} 
\crefformat{subsection}{\S~#2#1#3}
\crefformat{subsubsection}{\S~#2#1#3}

\makeatletter
\newcommand\footnoteref[1]{\protected@xdef\@thefnmark{\ref{#1}}\@footnotemark}
\makeatother

\newcommand{\ours}{{G\&O}\xspace}

\newcommand*{\rom}[1]{\uppercase\expandafter{\romannumeral #1}}

\newcommandx{\chao}[2][1=]{\todo[linecolor=red,backgroundcolor=red!25,bordercolor=red,#1]{#2}}
\newcommandx{\yli}[2][1=]{\todo[linecolor=blue,backgroundcolor=blue!25,bordercolor=blue,#1]{#2}}

\newcommand{\etc}{\emph{etc.}\xspace}
\newcommand{\eg}{\emph{e.g.}\xspace} 

\newcommand{\fone}{F\textsubscript{1}\xspace}

%% file: sections/0.abs.tex
\begin{abstract}
  Large language models (LLMs) have demonstrated impressive abilities in generating unstructured natural language according to instructions.
  However, their performance can be inconsistent when tasked with producing text that adheres to specific structured formats, which is crucial in applications like named entity recognition (NER) or relation extraction (RE).
  To address this issue, this paper introduces an efficient method, \ours, to enhance their structured text generation capabilities.
  It breaks the generation into a two-step pipeline: initially, LLMs generate answers in natural language as intermediate responses.
  Subsequently, LLMs are asked to organize the output into the desired structure, using the intermediate responses as context.
  \ours effectively separates the generation of content from the structuring process, reducing the pressure of completing two orthogonal tasks simultaneously.
  Tested on zero-shot NER and RE, the results indicate a significant improvement in LLM performance with minimal additional efforts.
  This straightforward and adaptable prompting technique can also be combined with other strategies, like self-consistency, to further elevate LLM capabilities in various structured text generation tasks.
\end{abstract}

%% file: sections/1.intro.tex
\section{Introduction}
\label{sec:intro}

Information extraction (IE) is a critical task that involves retrieving specific information, such as named entities and relationships, from unstructured or semi-structured texts, and converting this information into a structured format \citep{Cowie.1996.IE, Li.2023.TrENC}.
Traditionally, IE models have relied heavily on fully supervised learning, necessitating extensive labeled datasets for training.
This approach not only demands significant human effort but also restricts the scope of extractable information to a limited set of predefined types, such as ``\texttt{person}'' for Named Entity Recognition (NER) and ``\texttt{born in}'' for Relation Extraction (RE).
This limitation is particularly prominent in specialized fields like materials science, where resources are scarce.
Earlier attempts to mitigate these challenges, such as weak supervision \citep{Ren.2020.Denoising, Liang.2020.BOND, Zhang.2021.WRENCH, Li.2022.Sparse-CHMM}, have introduced methods that utilize noisy heuristic labeling functions (LFs) to reduce the reliance on manually labeled data.
However, the effectiveness of these methods often hinges on the quality of the LFs, which is not always consistent.

The advent of Large Language Models (LLMs) like GPTs \citep{OpenAI.2022.chatgpt, OpenAI.2023.GPT4} has promoted an attention shift towards universal IE approaches.
These methods aim to extract a wide range of information without the need for task-specific labels.
Strategies include directly prompting LLMs with instructions for specific tasks \citep{Wang.2023.GPT-NER, Han.2023.is.IE.solved, Xie.2023.ZeroShotNER, Zhang.2023.aligning.instruction.tasks} or fine-tuning them on either true labels or pseudo labels generated by GPTs \citep{Wang.2023.InstructUIE, Zhou.2023.UniversalNER, Sainz.2023.GoLLIE, Zaratiana.2023.GLiNER, Jiao.2023.Instruct.and.Extract}.
Nonetheless, the inherent mismatch between the unstructured data language models are typically trained on and the structured output requirement presents a challenge.
Previous studies have employed specialized prompts to guide the model in generating structured outputs, such as lists \citep{Zhou.2023.UniversalNER} or tables \citep{Jiao.2023.Instruct.and.Extract}.
However, integrating task instructions with these organizational prompts has sometimes resulted in formatting issues or compromised IE performance.

\input{sections/figures/s2.pipeline.tex}

In response to these challenges, this paper introduces a simple but effective methodology, Generate and Organize (\ours), designed to enhance the capability of LLMs in performing structured zero-shot IE tasks, with a focus on NER and RE.
Our approach divides the generation into two distinct components: 1) generating IE responses in a free-form natural language format; followed by 2) structuring these responses into a predefined format.
Furthermore, we incorporate a clean-up component to eliminate any potential noise from the free-form responses before structuring.
Through extensive experimentation, we demonstrate that our method boosts zero-shot IE performance across various LLMs.
Additionally, we show that each component of our approach contributes to its overall effectiveness.
Beyond NER and RE, \ours is versatile enough to be integrated with other techniques, such as self-consistency \citep{Wang.2023.Self-Consistency}, and can be applied to a broad spectrum of tasks requiring formatted outputs.
To support further research, we have made our methodology, including the code and experimental results, publicly available at \href{https://github.com/Yinghao-Li/GnO-IE}{https://github.com/Yinghao-Li/GnO-IE}.

%% file: sections/figures/s2.pipeline.tex
\begin{figure*}[!t]
    \centerline{\includegraphics[width = 0.9\textwidth]{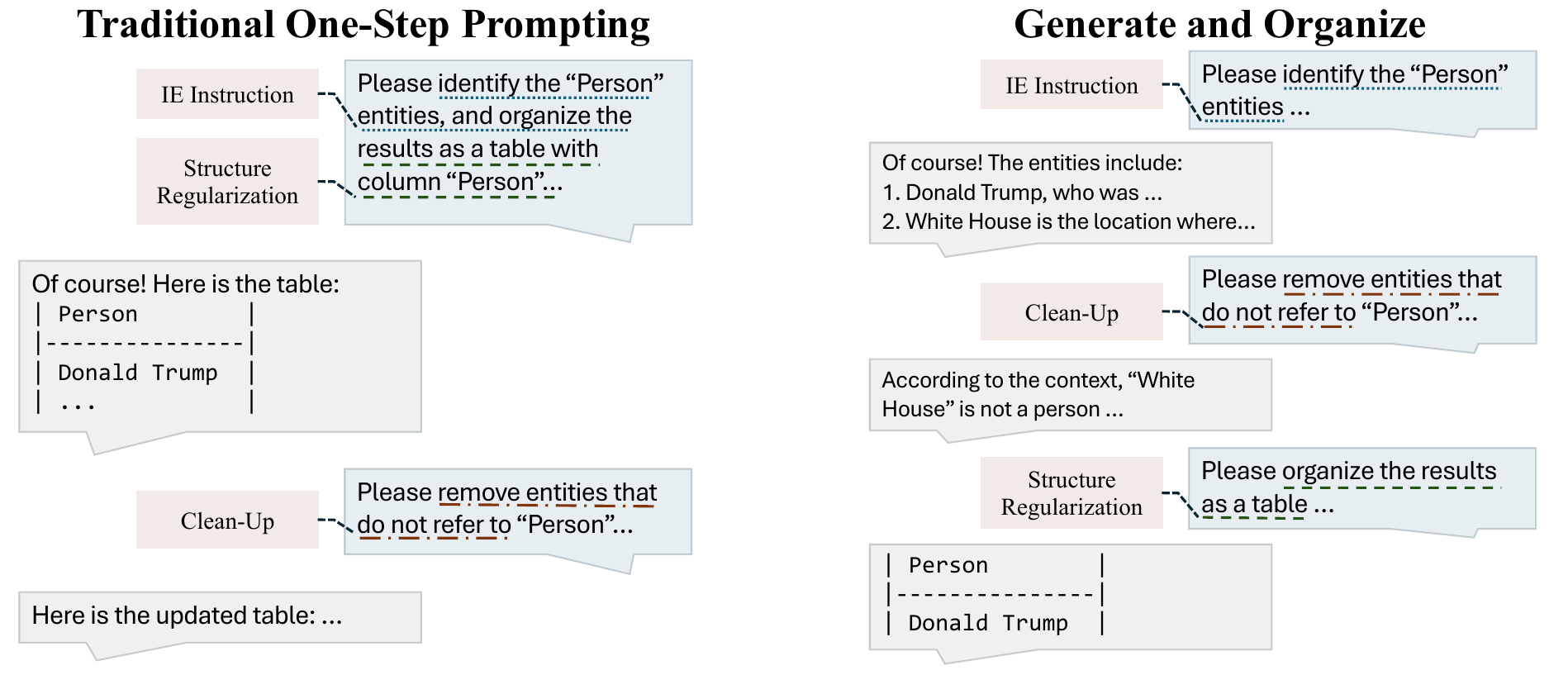}}
    \caption{
        The pipeline of \ours for NER, compared with Traditional One-Step prompting methods.
    }
    \label{fig:pipeline}
\end{figure*}

%% file: sections/8.related.tex
\section{Related Works}
\label{sec:related.works}

In the landscape of supervised neural networks, researchers are actively seeking methods to reduce reliance on labeled data for Information Extraction (IE) tasks, acknowledging the effort and limitations associated with manually labeled data.
A prominent approach includes the development of weak and distant supervision techniques \citep{Liang.2020.BOND, Lan.2020.Learning.to.Contextually.Aggregate, Lison.2020.NER.weak.supervision, Lison.2021.skweak, Zhang.2021.WRENCH, Yu.2021.COSINE, Li.2021.CHMM, Li.2022.Sparse-CHMM, Chen.2023.Neural.Hidden.CRF}.
These methods aim to lessen the annotation workload through the use of heuristic labeling functions (LFs).
These functions, whether singular \citep{Ren.2018.Learning.to.Reweight.Examples, Liang.2020.BOND, Yu.2021.COSINE} or multiple \citep{Lison.2020.NER.weak.supervision, Li.2021.CHMM, Li.2022.Sparse-CHMM, Chen.2023.Neural.Hidden.CRF, Lang.2022.Training.Subset.Selection}, are designed to generate noisy labels for unlabeled data.
Subsequently, models are trained to refine and amalgamate these labels for improved prediction accuracy.
However, some critics argue that the efficacy of single-LF methods is highly dependent on the quality of the clean validation set \citep{Zhu.2023.Weaker.Than.You.Think}, and the creation of multiple LFs can be a labor-intensive process \citep{Safranchik.2020.Weakly.Supervised.Sequence.Tagging, Lison.2021.skweak}.

Another research trajectory involves few-shot and zero-shot learning techniques \citep{Han.2018.FewRel, Han.2019.OpenNRE, Baldini.2019.Matching.the.Blanks, Yang.2020.Structured.Nearest.Neighbor, Ma.2022.Decomposed.Meta-Learning, Li.2023.TypeAwareNER}. These methods are directed towards adapting IE models to new domains using minimal labeled examples.
In line with the rapid advancements in Large Language Models (LLMs), some studies have explored directly prompting these models for open-type IE tasks \citep{Wang.2023.GPT-NER, Han.2023.is.IE.solved, Xie.2023.ZeroShotNER, Guo.2023.Retrieval.Augmented.Code.Generation}.
Additionally, there is an emerging focus on fine-tuning generative LLMs to better align with specific prompts or task formats \citep{Zhou.2023.UniversalNER, Zhang.2023.aligning.instruction.tasks, Sainz.2023.GoLLIE, Jiao.2023.Instruct.and.Extract}.
Nonetheless, these studies have overlooked the issue of LLMs' suboptimal performance in structured prediction tasks when using mixed prompts, which is the central topic of our research.

%% file: sections/4.method.tex
\section{Generate and Organize}
\label{sec:setup}

To enhance the capability of LLMs on zero-shot IE tasks that necessitate structured outputs, our prompting pipeline integrates three key components, as depicted in Figure~\ref{fig:pipeline}:
1) \textit{free-form response generation}, which prompts LLMs to identify the required information from the provided context without imposing any syntactic or structural constraints on the result;
2) \textit{answer clean-up}, tailored to the specific task at hand, filters out extraneous information to maintain the integrity of the final structured output; and
3) \textit{structure organization}, which is responsible for transforming the unstructured responses into organized formats, such as Markdown tables or lists, based on the LLMs' prior responses within the conversation history.
In addition, we add zero-shot CoT \citep{Kojima.2022.Zero.Shot.CoT} to further improve the IE performance.

\input{sections/figures/s5.ner.case.study.tex}

Although our modification appears minor compared to traditional IE prompts that combine components 1 and 3, it enhances alignment with the inherent semantic progression of natural language, and yields responses that are both more coherent and informative, according to the theory of \citet{Xie.2022.in.context.learning}.
In addition, clean-up also plays a crucial role.
As illustrated in Figure~\ref{fig:ner.case.study}, our empirical analysis reveals that while models efficiently identify relevant entities or relationships, they often include unrelated information that does not pertain to the requested types.
This phenomenon largely stems from the models' training to be ``helpful'' through RLHF \citep{Ouyang.2022.follow.instructions}.
Despite the identification of irrelevant entities, their presence complicates the task of formatting the useful information during the structure organization phase.
Hence, the clean-up phase is crucial to ensuring that the output is concise and focused solely on the entities of interest.
In the final step, we opt for Markdown tables as the structured format due to their prevalence in LLM training datasets and to maintain consistency with the RE pipeline.

%% file: sections/figures/s5.ner.case.study.tex
\begin{figure}[!t]
    \centerline{\includegraphics[width = \linewidth]{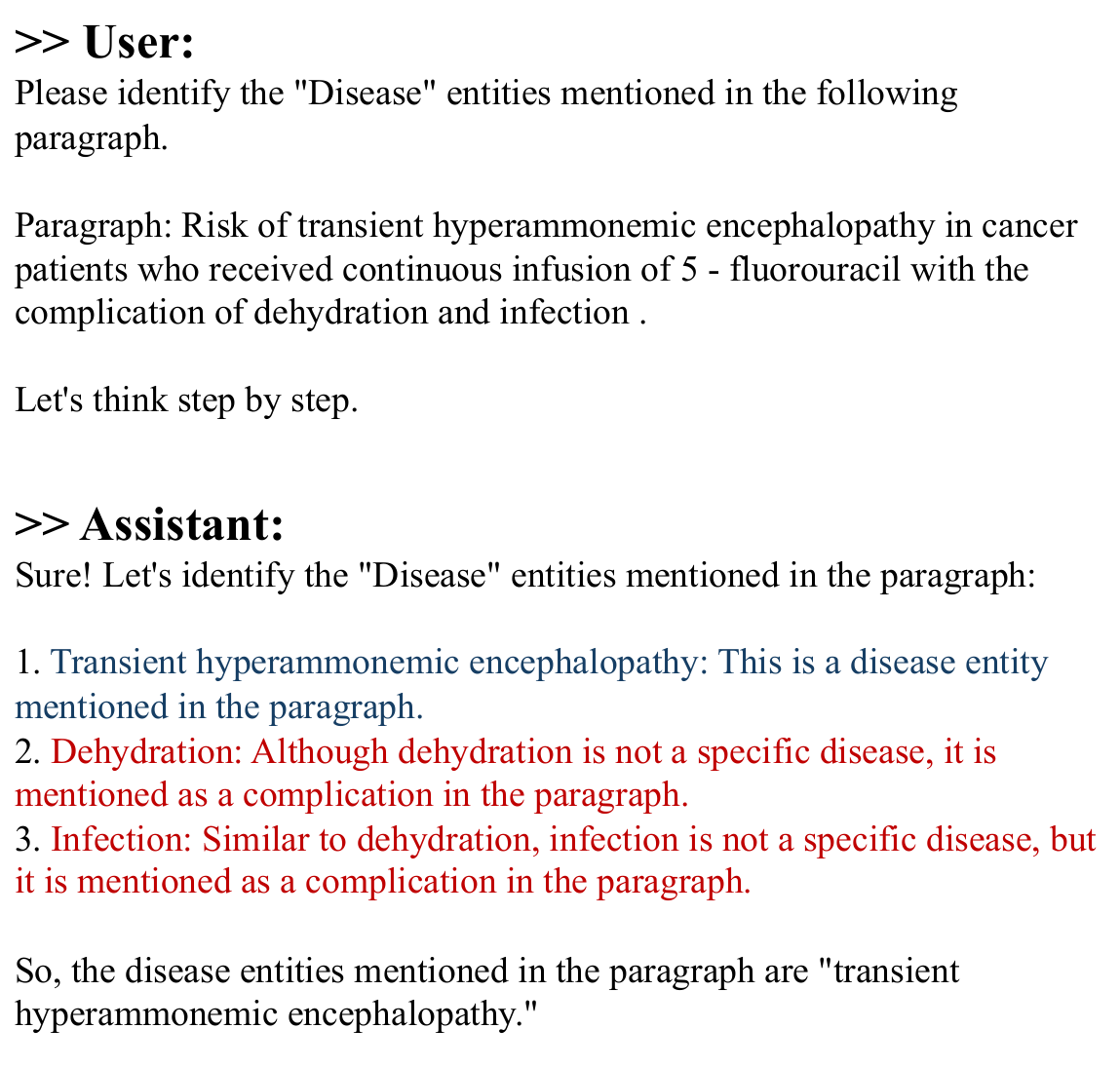}}
    \caption{
        GPT-3.5's natural language responses tend to include irrelevant entities (marked by red).
        Although clearly explained, irrelevant terms still pose a difficulty for GPT-3.5 during format organization.
    }
    \label{fig:ner.case.study}
\end{figure}

%% file: sections/5.results.tex
\input{sections/tables/s1.ner.f1.tex}

\section{Experiment and Discussion}
\label{sec:experiment}


\subsection{Named Entity Recognition}
\label{subsec:ner.result}

\paragraph{Datasets}


Our research utilizes diverse NER datasets , including CoNLL 2003 \citep{Sang.2003.CoNLL} from the general domain, NCBI Disease \citep{Dogan.2014.NCBIDisease} and BC5CDR \citep{Li.2016.BC5CDR} from the biomedical sector, and PolyIE \citep{Cheung.2023.PolyIE} from the field of materials science.
Please refer to \cref{subsec:setup.datasets} for statistics and details on data processing.

\paragraph{Baselines}

A fundamental baseline of \ours is \textbf{One-Step} prompting, which consolidates identification and organization into a single prompt.
While there are variations in implementation, this method is currently the dominant approach among LLMs for tasks demanding structured outputs.
We also consider another straightforward benchmark termed All-Entity-in-One (\textbf{AEiO}), which instructs the model to concurrently identify entities of various types, \eg, ``Identify \texttt{person}, \texttt{location}, and \texttt{organization} entities within the given paragraph''.
Note that One-Step also incorporates a clean-up phase, and AEiO differs from \ours primarily in the number of entity types it addresses.
Please refer to appendix~\ref{subsec:ner.setup} and \ref{subsec:re.setup} for details.

\paragraph{Metrics and Evaluation}


We employ micro-averaged precision, recall, and \fone score as metrics.
However, the strict span-level matching criterion disproportionately penalizes predictions that overlap with the ground truth without matching exactly, as observed by \citet{Zhou.2023.UniversalNER}.
Therefore, we use both \textbf{partial} and \textbf{full} matching scores.
The former acknowledges overlapping spans as true positives, whereas the latter demands identical entities.
Please refer to \cref{subsec:evalution.details} for more details.

\input{sections/figures/s5.ner.precision.recall.tex}

\paragraph{Main Results}

As our main NER results, Table~\ref{tb:ner.result} presents the \fone scores achieved by GPT-3.5 5 using various prompting strategies.
The effectiveness of \ours is evident when compared against the One-Step approach, where \ours-NER is superior in nearly all datasets under both partial and full matching criteria.
On average, the separation of task instruction and organization prompts yields a $15.8$\% increase in partial-match \fone and a $12.1$\% improvement in full.
Furthermore, the comparison with the AEiO baseline highlights the benefits of entity-specific instructions on information extraction performance.
Analysis of Figure~\ref{fig:ner.precision.recall} reveals that \ours-NER consistently boosts precision relative to the One-Step method without significantly affecting recall on average.
It shows that the opportunity to explain its results, facilitated by the CoT process, encourages GPT-3.5 to produce more precise final outputs through self-verification and clean-up.

\paragraph{Ablation Studies}


To assess the contribution of key elements within our approach, we conduct two straightforward ablation studies: excluding the CoT prompting and omitting the cleanup process.
Results in Table~\ref{tb:ner.result} show that the former leads to a $11.87$\% partial \fone drop on average and the latter $2.11$\%.
On a dataset-specific basis, these features exhibit minimal or even adverse effects on the CoNLL 2003 dataset.
However, they play a pivotal role in enhancing performance on scientific datasets.
As discussed in \cref{sec:setup}, LLMs are prone to integrating discussions about irrelevant scientific terms in their responses, a tendency less prevalent with general entities such as person names in the CoNLL dataset.
Moreover, the encouragement for models to articulate responses in natural language proves more advantageous for scientific datasets, where entities tend to be more complex and varied.

\paragraph{Resolving Entity Type Conflict}


Given \ours-NER processes each entity type separately, a notable challenge is the potential for a single entity span to be categorized under multiple types.
To mitigate this issue, we implement two strategies:
1) Conflict Resolution (\textbf{CR}), prompts LLMs to resolve any conflict of entity types as it arise; and
2) BERT Fine-Tuning (\textbf{FT}), which entails the fine-tuning of a pre-trained Transformer encoder \citep{Delvin.2019.BERT} using pseudo labels generated by GPT.
Detailed setup is provided in \cref{subsec:ner.setup}.
As indicated in Table~\ref{tb:ner.result}, both approaches enhance the overall effectiveness of \ours-NER, with FT being superior.
FT not only addresses the type conflict issue but also acts as a filter that discerns high-level entity patterns from the pseudo labels.
This process effectively refines the GPT-generated outputs by eliminating random inaccuracies.

\input{sections/figures/s5.ner.llms.tex}

\paragraph{Other LLMs}


In exploring the adaptability of our approach with various LLMs, we extended \ours to \num{4} open-source LLMs, including Llama 2 7B/70B \citep{Touvron.2023.Llama2}, Mistral 7B \citep{Jiang.2023.Mistral7B}, and Mixtral 8x7B \citep{Jiang.2024.Mixtral}, specifically their chat/instruct variants.
As depicted in Figure~\ref{fig:ner.llms}, the impact of \ours is less pronounced with these LLMs compared to GPT-3.5, indicating a dependency on the models' capacity for reasoning and following instructions.
Notably, Llama 2 models rarely produce explanations for their outputs, which renders \ours virtually equivalent to One-Step prompting, albeit less robust due to an increased likelihood of error propagation.
Conversely, \ours effectively encourages Mi[s/x]tral to provide detailed explanations in natural language, achieving a more consistent enhancement over the One-Step approach.
It can be concluded that \ours is more suited for LLMs that are designed with a focus on reasoning abilities and the capacity to engage in multi-round conversations.

\input{sections/figures/s5.re.gpt.tex}

\subsection{Relation Extraction}
\label{subsec:re.result}



In our study on RE, we evaluate \num{3} datasets: CoNLL 2004 \citep{Roth.2004.CoNLL-2004}, NYT \citep{Zeng.2018.NYT}, and PolyIE.
Our RE approach is end-to-end, predicting entities and their relations within the same iteration, which mirrors real-world scenarios and presents a greater challenge.
We primarily assess \ours-RE against the One-Step method using GPT-3.5, focusing on partial and full match precision, recall, and \fone scores.
Elaboration on \ours-RE is provided in \cref{subsec:re.setup}.

Figure~\ref{fig:re.gpt} illustrates that \ours enhances GPT-3.5's performance on RE tasks, registering an average \fone score improvement of $28.5$\% for partial matches and $7.6$\% for full matches.
Notably, across both NER and RE tasks, enhancements in partial matches consistently surpass those in full matches.
Analysis of LLM responses reveals that \ours tends to produce longer entity descriptions that incorporate attributes and modifiers not always present in the original annotations.
When applied to other LLMs, \ours-RE maintains consistent performance boosts, underscoring its versatility and applicability across diverse IE tasks.


%% file: sections/tables/s1.ner.f1.tex
\begin{table*}[t!] \small
  \centering
  \begin{tabular}{lcccccccc|cc}
    \toprule
    & \multicolumn{2}{c}{\textbf{CoNLL 2003}} & \multicolumn{2}{c}{\textbf{BC5CDR}} & \multicolumn{2}{c}{\textbf{NCBI Disease}} & \multicolumn{2}{c|}{\textbf{PolyIE}} & \multicolumn{2}{c}{\textbf{Macro Average}}\\ 
    \cmidrule(lr){2-3} \cmidrule(lr){4-5} \cmidrule(lr){6-7} \cmidrule(lr){8-9} \cmidrule(lr){10-11}
    & Partial & Full & Partial & Full & Partial & Full & Partial & Full & Partial & Full \\ 
    \midrule
    AEiO & 0.5370  & 0.4965  & 0.6199  & 0.5058  & - & - & 0.1300  & 0.0935 & 0.4290 & 0.3653 \\
    One-Step & 0.4741  & 0.4477  & 0.7030  & 0.6041  & 0.6500  & 0.5131  & 0.4669 & 0.3207 & 0.5735 & 0.4714 \\
    \midrule
    \textbf{\ours-NER} & 0.6569  & 0.6192  & 0.7610  & 0.6079  & 0.6935  & 0.5047  & 0.5449  & 0.3823 & 0.6641 & 0.5285 \\
    \midrule
    \quad $-$ CoT & 0.6572 & 0.6079 & 0.6634 & 0.5544 & 0.5653 & 0.4059 & 0.4551 & 0.3068 & 0.5853 & 0.4688 \\
    \quad $-$ clean-up & 0.7003  & 0.6436  & 0.7421  & 0.5861  & 0.6475  & 0.4541  & 0.5103  & 0.3421 & 0.6501 & 0.5065 \\
    \quad $+$ CR & 0.6775  & 0.6394  & 0.7724  & 0.6186  & - & - & 0.6011  & 0.4236 & 0.6837 & 0.5605 \\
    \quad $+$ FT & 0.7175  & 0.6800  & 0.7949  & 0.6838  & 0.7703  & 0.5507  & 0.7608 & 0.5533 & 0.7609 & 0.6170 \\
    \bottomrule
  \end{tabular}
  \caption{
    The \fone scores of GPT-3.5 on the NER datasets with different prompting strategies.
    ``Partial'' and ``Full'' refer to the partial and full matching criteria;
    ``$+$'' and ``$-$'' indicate the addition and removal of the corresponding components.
    ``CR'' stands for Conflict Resolution, and ``FT'' for BERT fine-tuning.
    AEiO and CR not applicable on NCBI Disease as it has only one entity type.
  }
  \label{tb:ner.result}
\end{table*}

%% file: sections/figures/s5.ner.precision.recall.tex
\begin{figure}[!t]
    \centerline{\includegraphics[width = \linewidth]{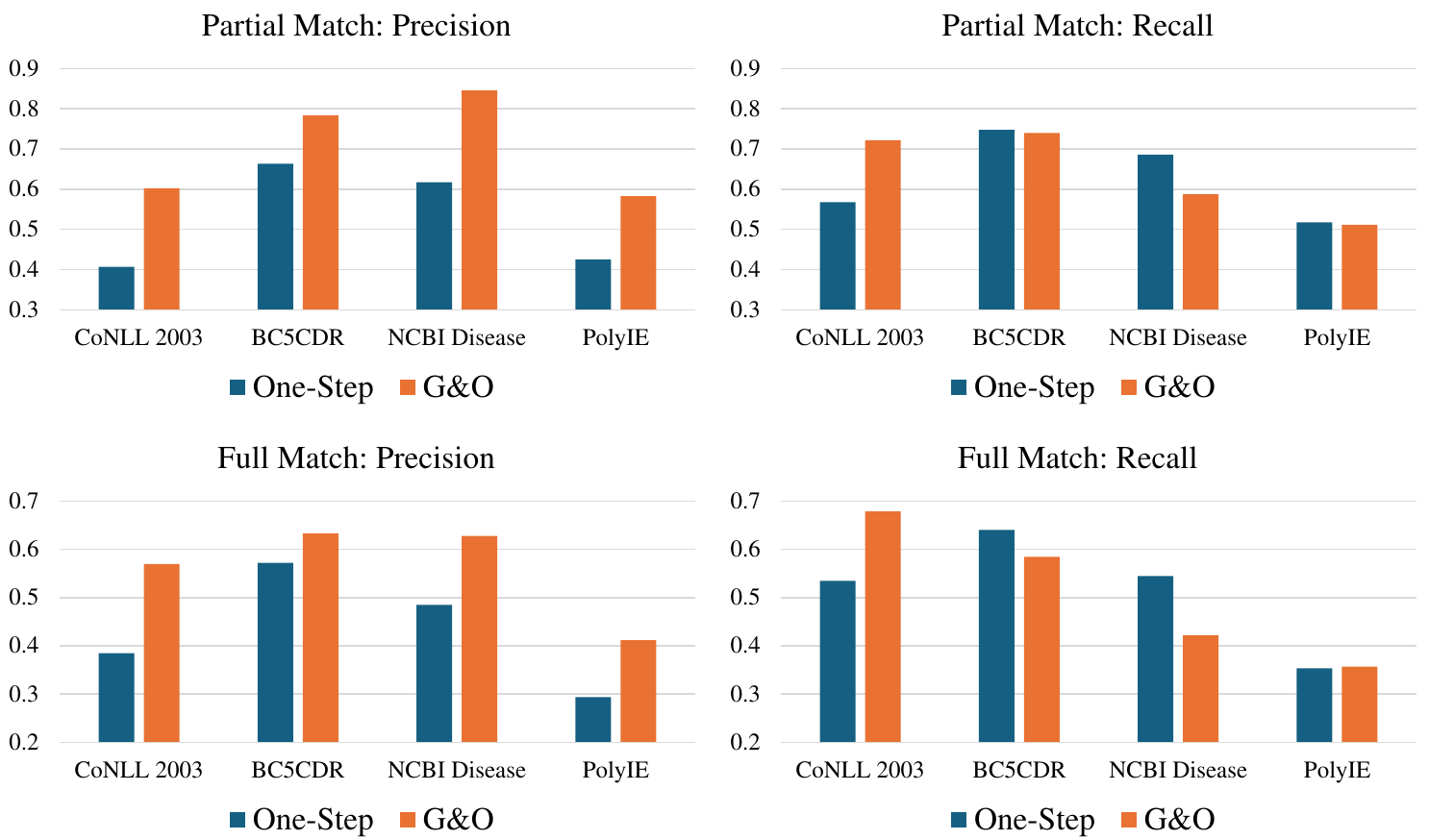}}
    \caption{
        Comparing the precision and recall of \ours-NER with One-Step on NER datasets.
    }
    \label{fig:ner.precision.recall}
\end{figure}

%% file: sections/figures/s5.ner.llms.tex
\begin{figure}[!t]
    \centerline{\includegraphics[width = \linewidth]{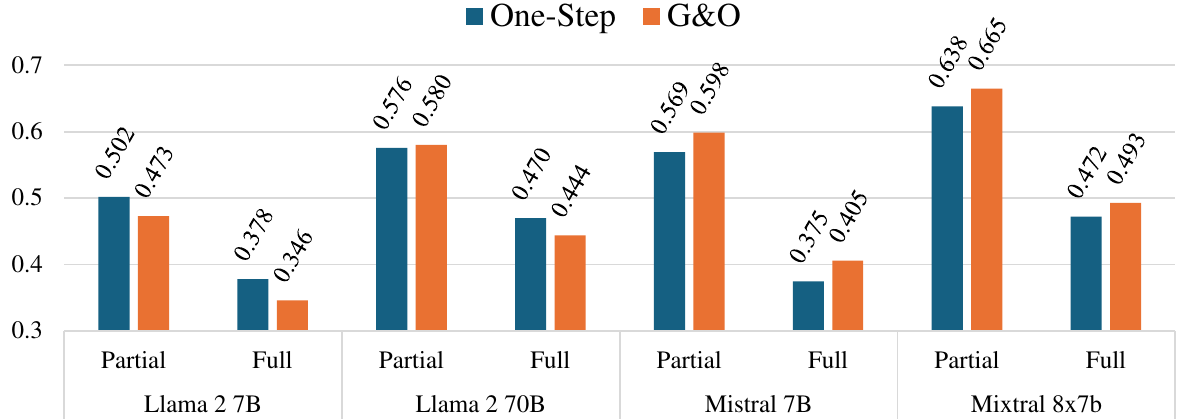}}
    \caption{
        \fone scores of differnt LMs with \ours and One-Step promptings, macro-averaged on the all datasets.
    }
    \label{fig:ner.llms}
\end{figure}

%% file: sections/figures/s5.re.gpt.tex
\begin{figure}[!t]
    \centerline{\includegraphics[width = \linewidth]{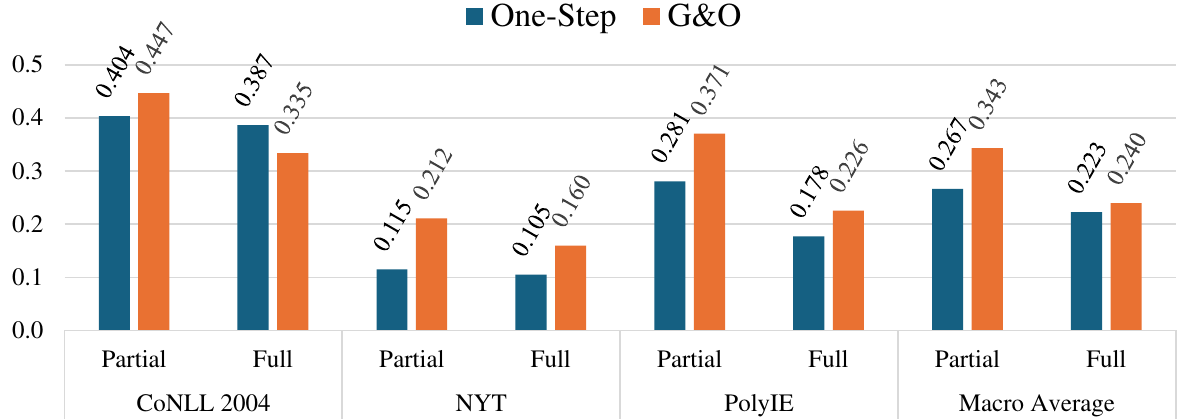}}
    \caption{
        The \fone scores of GPT-3.5 with different prompting approaches on RE datasets.
    }
    \label{fig:re.gpt}
\end{figure}

%% file: sections/6.conclusion.tex
\section{Conclusion}
\label{sec:conclusion}

In this paper, we propose a simple yet effective approach---\ours---to improve structured prediction from LLMs for IE tasks.
Different from conventional prompting approaches, \ours separates the identification and formatting steps into two stages, which allows the model to focus on each step independently and facilitates the generation of organized results.
Tested under the zero-shot IE settings with GPT 3.5, this simple adjustment brings significant performance gains, demonstrating the effectiveness of \ours.
The improvement is further validated by ablation studies and the generalizability of \ours to other LLMs, which can be further improved by resolving prediction conflicts using both prompting and fine-tuning methods.
We hope our work can bring insights and inspirations for further research on structured prediction from LLMs and contribute to the development of more effective and interpretable IE systems.

%% file: sections/7.limit.tex
\section*{Limitations}


Given the limitations of our computational resources, our evaluation was conducted on a select number of datasets and tasks.
While these experiments effectively illustrate the efficacy of \ours, we recognize that incorporating additional datasets and tasks could enhance the robustness of our conclusions.
Moreover, we exclusively utilized Markdown tables for structuring the final output, aiming for consistency across NER and RE tasks.
However, we did not investigate alternative formats such as lists or JSON, which are potentially more compatible with GPT models.
Such investigations are earmarked for future research endeavors.

Another intriguing aspect, not covered in this study, is the potential for fine-tuning open-source LLMs to align with our prompting format through methods like supervised fine-tuning and reinforcement learning.
We hypothesize that such an approach could significantly augment the zero-shot capabilities of universal IE tasks.
Addressing these areas of interest remains a key objective for our subsequent research efforts.

%% file: sections/9.ack.tex
\section*{Acknowledgments}

This work was supported in part by NSF IIS-2008334, IIS-2144338, and ONR MURI N00014-17-1-2656.

%% file: appendix/a2.setup.tex
\section{Detailed Experiment Setup}
\label{a2.setup}

\subsection{Models}
\label{a2.setup.models}

Our research primarily centers on GPT-3.5, particularly the \texttt{gpt-3.5-turbo-0613} version, as documented by \citep{OpenAI.2022.chatgpt}.\footnote{\href{https://platform.openai.com/docs/models/gpt-3-5-turbo}{platform.openai.com/docs/models/gpt-3-5-turbo}}.
Despite not being the most current iteration, we have opted to continue using this version to ensure the continuity of our experimental work.
In terms of open-source LLMs, our selection includes
Llama 2 7B (\texttt{Llama-2-7b-chat-hf},\footnote{\href{https://huggingface.co/meta-llama/Llama-2-7b-chat-hf}{huggingface.co/meta-llama/Llama-2-7b-chat-hf}} \citealp{Touvron.2023.Llama2}),
Llama 2 70B (\texttt{Llama-2-70b-chat-hf},\footnote{\href{https://huggingface.co/meta-llama/Llama-2-70b-chat-hf}{huggingface.co/meta-llama/Llama-2-70b-chat-hf}} \citealp{Touvron.2023.Llama2}),
Mistral 7B (\texttt{Mistral-7B-Instruct-v0.2},\footnote{\href{https://huggingface.co/mistralai/Mistral-7B-Instruct-v0.2}{huggingface.co/mistralai/Mistral-7B-Instruct-v0.2}} \citealp{Jiang.2023.Mistral7B}),
and Mixtral 8x7B (\texttt{Mixtral-8x7B-Instruct-v0.1},\footnote{\href{https://huggingface.co/mistralai/Mixtral-8x7B-Instruct-v0.1}{huggingface.co/mistralai/Mixtral-8x7B-Instruct-v0.1}} \citealp{Jiang.2024.Mixtral}).
Our experimental procedure involves only forward inference without any model fine-tuning.
The inference process for GPT 3.5 utilizes the OpenAI API via Azure, whereas the open-source LLMs are run using HuggingFace Transformers library \citep{Wolf.2019.HuggingFace} and vllm \citep{Kwon.2023.PagedAttention}.
The deployment of Llama 2 7B and Mistral 7B is each on an individual NVIDIA A100 80G GPU, while Mixtral 8x7B and Llama 2 70B are deployed on two GPUs each.

\input{appendix/tables/a3.ner.data.tex}

\subsection{Datasets}
\label{subsec:setup.datasets}

In the NER task, we incorporate datasets from several sources: CoNLL 2003 \citep{Sang.2003.CoNLL}, NCBI Disease \citep{Dogan.2014.NCBIDisease}, BC5CDR \citep{Li.2016.BC5CDR}, and PolyIE \citep{Cheung.2023.PolyIE}.
The CoNLL 2003, NCBI Disease, and BC5CDR datasets are obtained as prepared by \citet{Wang.2023.InstructUIE}, while the PolyIE dataset is sourced directly from \citet{Cheung.2023.PolyIE}.
We apply minor modifications to these datasets to tailor them to our study's needs.
Specifically, for CoNLL 2003, we remove the ``\texttt{MISC}'' entities due to their lack of informativeness and rare usage in practical scenarios.
In the case of PolyIE, ``\texttt{Condition}'' entities are excluded owing to the ambiguity surrounding their definition.
For BC5CDR, we limit our testing to \num{1000} randomly selected samples to minimize computational demands.
Preliminary experiments indicate that model performance on these subsets aligns with results obtained using the full datasets.
Furthermore, we adapt the input context for PolyIE to account for differences in the capabilities of GPT-3.5 and other language models.
For GPT-3.5, we input entire paragraphs, whereas for other LLMs, we broke the paragraphs into sentences and processed them individually, acknowledging their constrained history comprehension and memory abilities.
We report the performance on the test set of each dataset, and the detailed statistics are shown in Table~\ref{tb:ner.dataset}.

\input{appendix/tables/a3.re.data.tex}

In terms of RE, we utilize CoNLL 2004 \citep{Roth.2004.CoNLL-2004}, NYT \citep{Zeng.2018.NYT}, and PolyIE \citep{Cheung.2023.PolyIE}.
Similar to NER, we obtain CoNLL 2004 and NYT from \citet{Wang.2023.InstructUIE} and PolyIE from \citet{Cheung.2023.PolyIE}.
The infrequent relation types are also removed from the datasets to ensure a more focused and affordable evaluation.
For PolyIE, we only keep the ``\texttt{Material}-\texttt{Property}-\texttt{Value}'' relations and test all models using paragraph-based instances as the relations usually span multiple sentences.
The statistics of the RE datasets are shown in Table~\ref{tb:re.dataset}.

\subsection{NER Implementation Details}
\label{subsec:ner.setup}

\paragraph{\ours and Baselines}
Our experiments focus on revealing the difference between the zero-shot performance of LLMs when prompted with \ours and One-Step.
Therefore, we use the same prompt for both methods to ensure a fair comparison, except for the position of structure organization instruction.
Specifically, for the NER task, \ours-NER prompts are designed as Listing~\ref{ls:gno.ner.prompts}:

\begin{lstlisting}[caption=\ours-NER prompts., label=ls:gno.ner.prompts]
>> SYSTEM PROMPT
You are a knowledgeable assistant specialized in recognizing and understanding named entities and their interrelations. If requested to organize information in tabular format, you are adept at filtering and presenting only the relevant and valid results. You will exclude any results that are not pertinent or are inaccurate from the table according to the discussion history.

>> USER PROMPT  # Step 1. Free-form response generation
Please identify the "<ENTITY TYPE>" entities in the following paragraph.

Paragraph: <PARAGRAPH>

# optional zero-shot CoT prompt
Let's think step by step.

>> ASSISTANT ANSWER

# varies from case to case, omitted

>> USER PROMPT  # Step 2. Clean-up (optional)

Please remove irrelevant entities and only keep the entities that clearly refer to <ENTITY TYPE>.

>> ASSISTANT ANSWER

# varies from case to case, omitted

>> USER PROMPT  # Step 3. Structure organization

Please present the valid entities as a Markdown table with one column "<ENTITY TYPE>".

Make sure to present the entities precisely in the same words as in the original paragraph.

>> ASSISTANT ANSWER

# varies from case to case, omitted

\end{lstlisting}

In the prompts, the ``entity types'' are rephrased so that they are more comprehensible to the models.
For example, ``\texttt{PER}'' is rephrased as ``\texttt{person}''; ``\texttt{CN}'' is rephrased as ``\texttt{Material Name}'', \etc.
The prompting format is kept consistent across all models with only one difference: the system prompt is not applied to open-source LLMs.
Of course, the specific prompt string is adjusted to match each model's prompting style.
Similarly, the One-Step NER prompts are shown below:

\begin{lstlisting}[caption=One-Step NER prompts., label=ls:one-step.ner.prompts]
>> SYSTEM PROMPT
You are a knowledgeable assistant specialized in recognizing and understanding named entities and their interrelations. If requested to organize information in tabular format, you are adept at filtering and presenting only the relevant and valid results. You will exclude any results that are not pertinent or are inaccurate from the table according to the discussion history.

>> USER PROMPT  # Step 1. Free-form response generation
Please identify the "<ENTITY TYPE>" entities in the following paragraph and present the valid entities as a Markdown table with one column "<ENTITY TYPE>". Make sure to present the entities precisely in the same words as in the original paragraph.

Paragraph: <PARAGRAPH>

# optional zero-shot CoT prompt
Let's think step by step.

>> ASSISTANT ANSWER

# varies from case to case, omitted

>> USER PROMPT  # Step 2. Clean-up (optional)

Please remove irrelevant entities and only keep the entities that clearly refer to <ENTITY TYPE> and present the valid entities as a Markdown table with one column "<ENTITY TYPE>". Make sure to present the entities precisely in the same words as in the original paragraph.

>> ASSISTANT ANSWER

# varies from case to case, omitted
\end{lstlisting}

Notice that the term ``One-Step'' refers to the IE generation and structure organization being performed in a single step.
The pipeline could also contain a standalone clean-up step, which is set up as default in our experiments (as revealed in Listing~\ref{ls:one-step.ner.prompts}).

For both \ours and One-Step, we ask LLMs to generate the entities of one type at a time, which is most frequently adopted in previous works \citep{Wang.2023.InstructUIE,Xie.2023.ZeroShotNER}.
To validate the effectiveness of such an approach, we introduce AEiO, a method that generates all entities at once.
It is derived from \ours, with the only difference being the instruction to generate and organize all entities simultaneously.
The specific prompt for AEiO is shown in Listing~\ref{ls:aeio.ner.prompts}.

\begin{lstlisting}[caption=AEiO NER prompts., label=ls:aeio.ner.prompts]
>> SYSTEM PROMPT
You are a knowledgeable assistant specialized in recognizing and understanding named entities and their interrelations. If requested to organize information in tabular format, you are adept at filtering and presenting only the relevant and valid results. You will exclude any results that are not pertinent or are inaccurate from the table according to the discussion history.

>> USER PROMPT  # Step 1. Free-form response generation
Please identify the "<ENTITY TYPE 1, ENTITY TYPE 2, ..., ENTITY TYPE n>" entities in the following paragraph.

Paragraph: <PARAGRAPH>

# optional zero-shot CoT prompt
Let's think step by step.

>> ASSISTANT ANSWER

# varies from case to case, omitted

>> USER PROMPT  # Step 2. Clean-up (optional)

Please remove entities that do not clearly refer to any of the following entity types: "<ENTITY TYPE 1, ENTITY TYPE 2, ..., ENTITY TYPE n>".

>> ASSISTANT ANSWER

# varies from case to case, omitted

>> USER PROMPT  # Step 3. Structure organization

Please present the valid entities as a Markdown table with columns ["Entity", "Entity Type"].

Make sure to present the entities precisely in the same words as in the original paragraph and the Entity Types are in "<ENTITY TYPE 1, ENTITY TYPE 2, ..., ENTITY TYPE n>".

>> ASSISTANT ANSWER

# varies from case to case, omitted

\end{lstlisting}

As NCBI Disease contains only one entity type, AEiO is identical to \ours and is not applied.

\paragraph{Conflict Resolution}

As mentioned in \cref{subsec:ner.result}, we adopt two strategies to resolve the entity type conflict in NER results: Conflict Resolution (CR) and BERT Fine-Tuning (FT).
CR first identifies the conflicting entity spans and then prompts the model to resolve the conflict within the previous conflicting LLM outputs, also following the generation-organization prompting format.
An example of CR is shown in Listing~\ref{ls:cr}.

\begin{lstlisting}[caption=An example of conflict resolution on the CoNLL 2003 dataset., label=ls:cr]
>> SYSTEM PROMPTn
You are a helpful assistant who is good at identifying named entities and their relations.

>> USER PROMPT
According to the following paragraph, please identify and resolve the conflicts in the Named Entity Recognition (NER) results:

Paragraph: But China saw their luck desert them in the second match of the group , crashing to a surprise 2 - 0 defeat to newcomers Uzbekistan .

NER Response 1:
| Location    |
|-------------|
| China       |
| Uzbekistan  |

NER Response 2:
| Organization |
|--------------|
| Uzbekistan   |

Let's think step by step.

>> ASSISTANT ANSWER

Step 1: Identify the conflicts in the NER results.

In NER Response 1, both "China" and "Uzbekistan" are identified as locations, which is consistent with the context of the paragraph.

In NER Response 2, "Uzbekistan" is identified as an organization, which is not consistent with the context of the paragraph. "Uzbekistan" should be identified as a location.

Step 2: Resolve the conflicts in the NER results.

Based on the context of the paragraph, the resolved NER results should be:

Resolved NER Response 1:
| Location    |
|-------------|
| China       |
| Uzbekistan  |

Resolved NER Response 2:
| Location |
|--------------|
| Uzbekistan   |

Therefore, the conflicts in the NER results have been resolved.

>> USER PROMPT

Please present the entities with corrected entity types as a Markdown table with columns ["Entity", "Entity Type"]. Make sure the Entities are expressed in the same words as the original answers and the Entity Types in ["location", "organization", "person"].

>> ASSISTANT ANSWER

Here are the corrected entity types presented as a Markdown table:

|   Entity   | Entity Type |
|------------|-------------|
|   China    |   location  |
| Uzbekistan |   location  |

\end{lstlisting}

In the example, we directly take the conflicting LLM outputs (``NER Response $n$'') as the input context and ask the model to resolve the conflict.
We are aware that this may not be the most effective way to resolve conflicts, but it is the most straightforward to demonstrate the necessity of CR.

\paragraph{Supervised BERT Fine-Tuning}

Another strategy to resolve the entity type conflict and further boost the performance is to fine-tune a Transformer encoder model on LLM-generated content.
To achieve this goal, we first align the LLM-generated entities to the tokenized paragraph (which is the same as our evaluation process discussed in \cref{subsec:evalution.details}), generating a set of token-level labels.
Specifically, we use BIO2 tagging scheme, where ``B'' denotes the beginning of an entity, ``I'' denotes the inside of an entity, and ``O'' denotes the outside of an entity.
If a conflict occurs, we randomly select one of the conflicting entities to be the pseudo label.
With the training dataset established, we fully fine-tune DeBERTa V3 \citep{He.2023.DeBERTaV3},\footnote{\href{https://huggingface.co/microsoft/deberta-v3-base}{huggingface.co/microsoft/deberta-v3-base}} a state-of-the-art pre-trained Transformer encoder model, following the supervised learning paradigm with Cross Entropy loss.
One tricky part is that our setting is more similar to transductive learning, as the model is fine-tuned and evaluated on the same dataset, although the gold labels are different.
To prevent overfitting to pseudo labels, we apply a dropout rate of $0.3$ to both self-attention and feed-forward layers.
In addition, we use a relatively large learning rate of $1 \times 10^{-4}$ with AdamW optimizer \citep{Loshchilov.2019.AdamW} and a linear learning rate scheduler with warm-up ratio of $0.1$.
On all datasets, the model is updated for around $300$ \emph{steps}---roughly $6$ epochs for CoNLL 2003, $10$ epochs for NCBI Disease and BC5CDR, and $15$ epochs for PolyIE, with batch sizes adjusted from $16$ to $64$ accordingly.
We do not apply any early stopping strategy due to the lack of a reliable validation signal.
All experiments are conducted on a single NVIDIA A100 80G GPU with full-precision floating point numbers (float32), implemented with PyTorch \citep{Paszke.2019.PyTorch} and HuggingFace Transformers library \citep{Wolf.2019.HuggingFace}.
No factorization or efficient tuning approach is adopted.

\subsection{RE Implementation Details}
\label{subsec:re.setup}

RE poses a greater challenge than NER because it demands that the model not only discern entities within the text but also understand their contextual relationships in an end-to-end manner.
Many relation labels, such as ``\texttt{place lived}'' or ``texttt{location contains}'', present ambiguity that can be difficult for LLMs to comprehend.
To mitigate this, we tailor prompts for each type of relation to enhance the model's comprehension.
Specifically, we leverage GPT-4 \citep{OpenAI.2023.GPT4} with Web UI to craft prompts based on a simple slot-filling template designed for GPT-3.5, enabling it to recognize specific relations from the textual context.
An example of this process is provided in Listing~\ref{ls:re.prompt.gen}, showcasing we guide GPT-4 in generating relation extraction prompts for GPT-3.5.

\begin{lstlisting}[caption=An example of GPT-4's instruction to generate RE prompts for GPT 3.5., label=ls:re.prompt.gen]
>> USER PROMPT
Please rephrase the following prompt so that GPT 3.5 can detect the "location contains" relationships between Florida and Boca Raton:

Original Prompt: Please identify the "location contains" relationships between the "Location" and "Location" entities in the following paragraph. 

Paragraph: Graveside service Monday January 31 , 2:00 P.M. at Riverside Cemetery , Rochelle Park , N.J. Donations may be made to Hospice By The Sea , Boca Raton , Florida .

Let's thinks step by step.
\end{lstlisting}

Notice that the ``original prompt'' is simply modified from the template ``Please identify the <\texttt{relation type}> relationships between the <\texttt{head entity type}> and <\texttt{tail entity type}> entities in the following paragraph.''
These example paragraphs and labels used for prompt construction are drawn randomly from the training partition of datasets, ensuring no test data is exposed.

Another distinct aspect of RE, compared to NER, is the integration of the clean-up process into the structural organization phase, rather than treating it as a separate step.
For RE, we introduce an additional column during the structuring phase, so that the result table not only lists entities linked by the desired relation but also indicates the presence of that relation.
In the post-processing stage, any entity pairs without a confirmed relation in this additional column are excluded.
This integrated approach has proven more effective in preliminary tests than the isolated clean-up process traditionally used in NER.
Examples of \ours-RE in Listing~\ref{ls:gno.re} conversation pipeline on the CoNLL 2004 dataset for the relation type ``\texttt{organization-based-in}'' illustrate this methodology.

\begin{lstlisting}[caption=An example \ours-RE., label=ls:gno.re]
>> SYSTEM PROMPT
You are a knowledgeable assistant specialized in recognizing and understanding named entities and their interrelations. When requested to organize information in tabular format, you are adept at filtering and presenting only the relevant and valid results. You will exclude any results that are not pertinent or are inaccurate from the table according to the discussion history.

>> USER PROMPT  # Step 1. Free-form response generation
Please analyze the given text to identify relationships where an organization is headquartered or primarily operates in a specific location. Look for patterns that indicate this type of relationship.

Paragraph: An art exhibit at the Hakawati Theatre in Arab east Jerusalem was a series of portraits of Palestinians killed in the rebellion .

Let's think step by step.

>> ASSISTANT ANSWER

# varies from case to case, omitted

>> USER PROMPT  # Step 2. Structure organization and clean-up

If exists, please present the valid relationships as a Markdown table with columns ["Organization", "Location", "Whether the Organization is based in the Location"]. Make sure the table items are from the original paragraph.

>> ASSISTANT ANSWER

# varies from case to case, omitted
\end{lstlisting}

Furthermore, the One-Step RE prompt merges the two steps of the aforementioned pipeline, simplifying the process into a single prompt, as shown in Listing~\ref{ls:os.re}.

\begin{lstlisting}[caption=An example One-Step prompting for RE., label=ls:os.re]
>> SYSTEM PROMPT
You are a knowledgeable assistant specialized in recognizing and understanding named entities and their interrelations. When requested to organize information in tabular format, you are adept at filtering and presenting only the relevant and valid results. You will exclude any results that are not pertinent or are inaccurate from the table according to the discussion history.

>> USER PROMPT
Please analyze the given text to identify relationships where an organization is headquartered or primarily operates in a specific location. Look for patterns that indicate this type of relationship. If exists, please present the valid relationships as a Markdown table with columns ["Organization", "Location", "Whether the Organization is based in the Location"]. Make sure the table items are from the original paragraph.

Paragraph: An art exhibit at the Hakawati Theatre in Arab east Jerusalem was a series of portraits of Palestinians killed in the rebellion .

Let's think step by step.

>> ASSISTANT ANSWER

# varies from case to case, omitted

\end{lstlisting}

For a comprehensive review of the prompts designed for each relation type, we refer readers to the meta files accompanying each dataset within our code repository.

\subsection{Post-Processing and Evaluation}
\label{subsec:evalution.details}

While extracting entities or relationships, the outputs from LLMs may not always align perfectly with the terminology or phrasing in the source text.
Issues such as extraneous or missing spaces, variations in tense, and unnecessary clarification of acronyms, are common, particularly for smaller models.
To address this, we employ a fuzzy matching algorithm using Python's difflib library\footnote{\href{https://docs.python.org/3/library/difflib.html}{docs.python.org/3/library/difflib.html}} to better correlate the LLM outputs with the original text.

Subsequently, we evaluate the model's precision, recall, and \fone score based on how well the predicted entity spans match the actual ground truth spans.
As discussed in \cref{subsec:ner.result}, our evaluation encompasses both full and partial match scores to provide a thorough assessment of model accuracy.
A full match necessitates complete agreement between the predicted and ground truth spans, aligning with traditional evaluation methods.
Partial matching, however, accounts for overlaps between predicted and actual spans, thus accommodating minor discrepancies.
For instance, in the sentence ``He's working for the White House'', a ground truth entity labeled ``White House $\rightarrow$ \texttt{Organization}'' and a predicted span ``the White House $\rightarrow$ \texttt{Organization}'' (with an added ``the'' in the span) would be acknowledged as a true positive prediction in a partial match scenario, but considered both a false positive and a false negative in a full match evaluation.
Conversely, labeling ``White House'' as a ``\texttt{Location}'' would be incorrect under both matching criteria.

For RE tasks, achieving a partial or full match on all entity spans in the relation group is required for a prediction to be considered correct in CoNLL 2004 and NYT.
In the PolyIE dataset, we adopt a more lenient approach, accepting a relation prediction as correct if at least one set of mapped entity spans within a paragraph corresponds to the ground truth.
This flexibility is due to the frequent mention of each entity within a relation group across the paragraph, owing to its length (Table~\ref{tb:re.dataset}).
Imposing a strict criterion for matching all entity spans in PolyIE could lead to counterintuitive evaluation outcomes.

%% file: appendix/tables/a3.ner.data.tex
\begin{table}[t!] \tiny
  \centering
  \begin{tabular}{lcccc}
    \toprule
    & \textbf{CoNLL 03} & \textbf{BC5CDR} & \textbf{NCBI} & \textbf{PolyIE} \\
    \midrule
    n-instance & 3,453 & 1,000 & 940 & 96 / 1,170 \\
    avg. l-text & 70 & 148 & 147 & 2,761 / 188 \\
    n-entity-type & 3 & 2 & 1 & 3 \\
    n-entity-mention & 4,945 & 2,074 & 957 & 4,803 \\
    \bottomrule
  \end{tabular}
  \caption{
    NER dataset statistics.
    ``avg. l-text'' denotes the average number of characters in each text instance.
    The statistics of PolyIE is shown as ``Paragraph-level / Sentence-level''.
  }
  \label{tb:ner.dataset}
\end{table}

%% file: appendix/tables/a3.re.data.tex
\begin{table}[t!] \small
  \centering
  \begin{tabular}{lccc}
    \toprule
    & \textbf{CoNLL 04} & \textbf{NYT}  & \textbf{PolyIE} \\
    \midrule
    n-instance & 288 & 369 & 96 \\
    avg. l-text & 159 & 199 & 2,761  \\
    n-relation-type & 5 & 7 & 1 \\
    n-ary-relations & 2 & 2 & 3 \\
    n-relation-mention & 422 & 265 & 527 \\
    \bottomrule
  \end{tabular}
  \caption{
    RE dataset statistics.
    ``n-ary-relations'' indicates the number of entities in a relation tuple (group).
  }
  \label{tb:re.dataset}
\end{table}

%% file: appendix/a3.results.tex
\input{appendix/tables/a1.ner.tex}

\input{appendix/tables/a2.re.tex}

\section{Complete Results}

Tables~\ref{tb:complete.ner.result} and \ref{tb:complete.re.result} present the complete results of our experiments on the NER and RE tasks, respectively.
Due to the limiation of computational resources, we do not conduct the full set of ablation studies on open-source LLMs or the RE task, and only validate our key point, \ours, by comparing it to the One-Step approach.
The key findings are presented using tables and figures in the main paper and will not be repeated here.

%% file: appendix/tables/a1.ner.tex
\begin{table*}[t!] \tiny
  \centering
  \begin{tabular}{clcccc}
    \toprule
    & & \multicolumn{2}{c}{\textbf{CoNLL 2003}} & \multicolumn{2}{c}{\textbf{BC5CDR}} \\
    \cmidrule(lr){3-4} \cmidrule(lr){5-6}
    & & Partial & Full & Partial & Full \\ 
    \midrule
    \multirow{7}[7]{*}{GPT-3.5}
    & AEiO & 0.5370 (0.6819 / 0.4429) & 0.4965 (0.6323 / 0.4088) & 0.6199 (0.8254 / 0.4963) & 0.5058 (0.6794 / 0.4028) \\
    & One-Step & 0.4741 (0.4070 / 0.5678) & 0.4477 (0.3850 / 0.5349) & 0.7030 (0.6632 / 0.7479) & 0.6041 (0.5720 / 0.6401) \\
    \cmidrule(lr){2-6}
    & \textbf{\ours-NER} & 0.6569 (0.6027 / 0.7219) & 0.6192 (0.5695 / 0.6784) & 0.7610 (0.7835 / 0.7398) & 0.6079 (0.6334 / 0.5845) \\
    \cmidrule(lr){2-6}
    & \quad $-$ CoT & 0.6572 (0.6648 / 0.6498) & 0.6079 (0.6185 / 0.5977) & 0.6634 (0.8001 / 0.5666) & 0.5544 (0.6776 / 0.4691) \\
    & \quad $-$ clean-up & 0.7003 (0.6482 / 0.7616) & 0.6436 (0.5992 / 0.6950) & 0.7421 (0.7153 / 0.7712) & 0.5861 (0.5699 / 0.6032) \\
    & \quad $+$ CR & 0.6775 (0.6386 / 0.7213) & 0.6394 (0.6043 / 0.6788) & 0.7724 (0.7954 / 0.7506) & 0.6186 (0.6447 / 0.5946) \\
    & \quad $+$ FT & 0.7175 (0.6496 / 0.8012) & 0.6800 (0.6161 / 0.7585) & 0.7949 (0.8166 / 0.7743) & 0.6838 (0.7068 / 0.6622) \\
    \midrule
    \multirow[c]{2}{*}{Llama 2 7B}
    & One-Step & 0.4237 (0.3234 / 0.6139) & 0.3929 (0.3005 / 0.5672) & 0.6426 (0.5766 / 0.7256) & 0.5087 (0.4608 / 0.5678) \\
    & \textbf{\ours-NER} & 0.4281 (0.3193 / 0.6495) & 0.3787 (0.2830 / 0.5725) & 0.6408 (0.6300 / 0.6520) & 0.5169 (0.5131 / 0.5207) \\
    \midrule
    \multirow[c]{2}{*}{Llama 2 70B}
    & One-Step & 0.4685 (0.3711 / 0.6353) & 0.4428 (0.3514 / 0.5983) & 0.7532 (0.7118 / 0.7997) & 0.6389 (0.6081 / 0.6730) \\
    & \textbf{\ours-NER} & 0.5476 (0.4490 / 0.7016) & 0.5128 (0.4213 / 0.6550) & 0.7260 (0.7251 / 0.7270) & 0.5792 (0.5849 / 0.5737) \\
    \midrule
    \multirow[c]{2}{*}{Mistral 7B}
    & One-Step & 0.4884 (0.3970 / 0.6344) & 0.4075 (0.3334 / 0.5240) & 0.7246 (0.7012 / 0.7496) & 0.5244 (0.5149 / 0.5342) \\
    & \textbf{\ours-NER} & 0.5693 (0.4963 / 0.6676) & 0.4963 (0.4349 / 0.5780) & 0.7318 (0.7967 / 0.6766) & 0.5300 (0.5870 / 0.4831) \\
    \midrule
    \multirow[c]{2}{*}{Mixtral 8x7B}
    & One-Step & 0.5827 (0.5023 / 0.6936) & 0.5213 (0.4517 / 0.6163) & 0.7815 (0.7574 / 0.8072) & 0.6038 (0.5923 / 0.6157) \\
    & \textbf{\ours-NER} & 0.6575 (0.6029 / 0.7230) & 0.5937 (0.5471 / 0.6489) & 0.7697 (0.7840 / 0.7560) & 0.6098 (0.6309 / 0.5902) \\

    \midrule
    \midrule
    & & \multicolumn{2}{c}{\textbf{NCBI Disease}} & \multicolumn{2}{c}{\textbf{PolyIE}} \\
    \cmidrule(lr){3-4} \cmidrule(lr){5-6}
    & & Partial & Full & Partial & Full \\ 
    \midrule
    \multirow{7}[7]{*}{GPT-3.5}
    & AEiO & - & - & 0.1300 (0.7440 / 0.0712) & 0.0935 (0.5383 / 0.0512) \\
    & One-Step & 0.6500 (0.6175 / 0.6860) & 0.5131 (0.4851 / 0.5445) & 0.4669 (0.4253 / 0.5177) & 0.3207 (0.2936 / 0.3533) \\
    \cmidrule(lr){2-6}
    & \textbf{\ours-NER} & 0.6935 (0.8458 / 0.5877) & 0.5047 (0.6278 / 0.4220) & 0.5449 (0.5830 / 0.5115) & 0.3823 (0.4117 / 0.3569) \\
    \cmidrule(lr){2-6}
    & \quad $-$ CoT & 0.5653 (0.8260 / 0.4297) & 0.4059 (0.6101 / 0.3041) & 0.4551 (0.4901 / 0.4249) & 0.3068 (0.3342 / 0.2836) \\
    & \quad $-$ clean-up & 0.6475 (0.6775 / 0.6200) & 0.4541 (0.4886 / 0.4242) & 0.5103 (0.5036 / 0.5173) & 0.3421 (0.3396 / 0.3448) \\
    & \quad $+$ CR & - & - & 0.6011 (0.6723 / 0.5572) & 0.4236 (0.4685 / 0.3866) \\
    & \quad $+$ FT & 0.7703 (0.8822 / 0.6837) & 0.5507 (0.6356 / 0.4859) & 0.7608 (0.7044 / 0.8270) & 0.5533 (0.5034 / 0.6141) \\
    \midrule
    \multirow[c]{2}{*}{Llama 2 7B}
    & One-Step & 0.5405 (0.4670 / 0.6416) & 0.3474 (0.3076 / 0.3992) & 0.3994 (0.3701 / 0.4338) & 0.2629 (0.2440 / 0.2849) \\
    & \textbf{\ours-NER} & 0.5342 (0.5481 / 0.5209) & 0.3000 (0.3163 / 0.2853) & 0.2881 (0.2898 / 0.2864) & 0.1874 (0.1886 / 0.1863) \\
    \midrule
    \multirow[c]{2}{*}{Llama 2 70B}
    & One-Step & 0.6390 (0.5910 / 0.6957) & 0.4608 (0.4340 / 0.4911) & 0.4421 (0.4307 / 0.4541) & 0.3355 (0.3255 / 0.3461) \\
    & \textbf{\ours-NER} & 0.5992 (0.6098 / 0.5890) & 0.3736 (0.3902 / 0.3584) & 0.4466 (0.4313 / 0.4630) & 0.3093 (0.2980 / 0.3215) \\
    \midrule
    \multirow[c]{2}{*}{Mistral 7B}
    & One-Step & 0.6715 (0.7211 / 0.6282) & 0.3842 (0.4306 / 0.3469) & 0.3923 (0.3357 / 0.4720) & 0.1821 (0.1586 / 0.2138) \\
    & \textbf{\ours-NER} & 0.6588 (0.7987 / 0.5606) & 0.3892 (0.4897 / 0.3229) & 0.4335 (0.4313 / 0.4357) & 0.2060 (0.2072 / 0.2048) \\
    \midrule
    \multirow[c]{2}{*}{Mixtral 8x7B}
    & One-Step & 0.6674 (0.6734 / 0.6615) & 0.4484 (0.4667 / 0.4316) & 0.5210 (0.4717 / 0.5817) & 0.3144 (0.2873 / 0.3472) \\
    & \textbf{\ours-NER} & 0.7049 (0.8629 / 0.5958) & 0.4413 (0.5613 / 0.3636) & 0.5275 (0.4609 / 0.6165) & 0.3268 (0.2880 / 0.3776) \\

    \bottomrule
  \end{tabular}
  \caption{
    The complete results of different LLMs on NER datasets, presented as ``\fone (precision / recall)''.
  }
  \label{tb:complete.ner.result}
\end{table*}

%% file: appendix/tables/a2.re.tex
\begin{table*}[t!] \tiny
  \centering
  \begin{tabular}{lcccccc}
    \toprule
    & \multicolumn{2}{c}{\textbf{CoNLL 2004}} & \multicolumn{2}{c}{\textbf{NYT}} & \multicolumn{2}{c}{\textbf{PolyIE}} \\
    \cmidrule(lr){2-3} \cmidrule(lr){4-5} \cmidrule(lr){6-7}
    & Partial & Full & Partial & Full & Partial & Full \\ 
    \midrule
    \multicolumn{7}{c}{\textbf{GPT-3.5}} \\ 
    \midrule
    One-Step & 0.404 (0.363 / 0.455) & 0.387 (0.261 / 0.324) &  0.116 (0.088 / 0.170) & 0.106 (0.080 / 0.155) & 0.281 (0.427 / 0.210) & 0.178 (0.256 / 0.136) \\
    \textbf{\ours-RE} & 0.447 (0.436 / 0.459) & 0.335 (0.333 / 0.337) & 0.212 (0.145 / 0.393) & 0.160 (0.110 / 0.298) & 0.371 (0.390 / 0.353) & 0.226 (0.221 / 0.231) \\
    \midrule
    \multicolumn{7}{c}{\textbf{Llama 2 70B}} \\ 
    \midrule
    One-Step & 0.334 (0.234 / 0.584) & 0.224 (0.159 / 0.378) & 0.182 (0.103 / 0.765) & 0.152 (0.086 / 0.645) & 0.358 (0.413 / 0.316) & 0.254 (0.277 / 0.234) \\
    \textbf{\ours-RE} & 0.361 (0.258 / 0.605) & 0.252 (0.184 / 0.404) & 0.191 (0.111 / 0.683) & 0.156 (0.091 / 0.559) & 0.371 (0.449 / 0.317) & 0.272 (0.313 / 0.241) \\
    \midrule
    \multicolumn{7}{c}{\textbf{Mixtral 8x7B}} \\ 
    \midrule
    One-Step & 0.423 (0.418 / 0.428) & 0.260 (0.264 / 0.257) & 0.261 (0.179 / 0.477) & 0.187 (0.128 / 0.343) & 0.242 (0.367 / 0.107) & 0.134 (0.193 / 0.102) \\
    \textbf{\ours-RE} & 0.441 (0.397 / 0.496) & 0.294 (0.270 / 0.323) & 0.237 (0.152 / 0.541) & 0.170 (0.109 / 0.389) & 0.364 (0.416 / 0.324) & 0.226 (0.240 / 0.213) \\
    \bottomrule
  \end{tabular}
  \caption{
    Comprehensive performance metrics of various LLMs on RE Datasets, expressed as \fone (precision / recall).
    Results from smaller-scale models, Llama 2 7B and Mistral 7B, are omitted due to their inability to produce valid responses in initial testing on NYT and PolyIE.
  }
  \label{tb:complete.re.result}
\end{table*}